\pdfoutput=1

\documentclass[runningheads]{llncs}
\usepackage{times}
\usepackage{graphicx}
\usepackage{mysubfigure}
\usepackage{amsmath,amssymb} 
\usepackage{color}
\usepackage[width=122mm,left=12mm,paperwidth=146mm,height=193mm,top=12mm,paperheight=217mm]{geometry}
\usepackage[pagebackref=true,breaklinks=true,letterpaper=true,colorlinks,bookmarks=false]{hyperref}

\begin{document}
\pagestyle{headings}
\mainmatter
\def\ECCV16SubNumber{}  

\title{Seeing Invisible Poses: Estimating 3D Body Pose from Egocentric Video}

\titlerunning{Seeing Invisible Poses: Estimating 3D Body Pose from Egocentric Video}

\authorrunning{H. Jiang and K. Grauman}

\author{Hao Jiang\inst{1} and Kristen Grauman\inst{2}}
\institute{Computer Science Department, Boston College, USA  
   \and Department of Computer Science, University of Texas at Austin, USA}

\maketitle

\begin{abstract}

Understanding the camera wearer's activity is central to egocentric vision, yet one key facet of that activity is inherently \emph{invisible} to the camera---the wearer's body pose.  Prior work focuses on estimating the pose of hands and arms when they come into view, but this 1) gives an incomplete view of the full body posture, and 2) prevents any pose estimate at all in many frames, since the hands are only visible in a fraction of daily life activities.  We propose to infer the ``invisible pose" of a person behind the egocentric camera.  Given a single video, our efficient learning-based approach returns the full body 3D joint positions for each frame.  Our method exploits cues from the dynamic motion signatures of the surrounding scene---which changes predictably as a function of body pose---as well as static scene structures that reveal the viewpoint (e.g., sitting vs. standing).  We further introduce a novel energy minimization scheme to infer the pose sequence.  It uses soft predictions of the poses per time instant together with a non-parametric model of human pose dynamics over longer windows.  Our method outperforms an array of possible alternatives, including deep learning approaches for direct pose regression from images.


\end{abstract}

%

\section{Introduction}


Wearable ``egocentric" cameras are steadily gaining traction---thanks not only to smaller devices, but also the increasing promise of vision and learning technology to transform applications.  Head- or chest-mounted cameras, initially perceived as the purview of hard-core life loggers,
are now valuable tools for many others.  Last year, President Obama authorized \$20M for law enforcement agencies across the US for purchasing bodycams in an effort to promote transparency with the public.  
Psychologists leverage wearable cameras on infants to gain insights into motor and linguistic development~\cite{smith}.  
In healthcare, 
egocentric vision could move daily-living activity monitoring required for motor rehabilitation from the hospital to the home~\cite{kopp,pirsiavash}.

For many applications, the important vision problems center around inferring the camera wearer's behavior, i.e., his activity and interactions with people and objects.  As such, the ability to infer the \emph{camera wearer's 3D body pose} is of great interest.  However, doing so is challenging because most body parts are invisible to the egocentric camera!

Existing work estimates a person's pose by analyzing the body parts visible in his first-person camera.  Naturally, this makes them restricted to the arms and hands~\cite{deva,kitani-iccv2013,kitani-cvpr2013} \cite{ren-gu,damen}.
However, from the view of a chest-mounted wide-angle camera,
arms and legs are often not visible in daily life activity.  
For example, in our ground truth videos in which people perform normal activities in
public places such as labs and offices, the chance to view
arms and legs is less than $10\%$.  
To estimate full body pose, one creative approach~\cite{insideout} is to fasten multiple cameras to all the person's joints, then use structure from motion (SfM) to localize the cameras and hence the joints.  However, this comes with the disadvantages of requiring 1) obtrusive multi-camera equipment not amenable to everyday casual use and 2) intensive computational requirements (hours to days of processing to infer pose for a minute of video~\cite{insideout}).

\begin{figure}
  \centering
  \subfigure[]{\includegraphics[width=0.176\textwidth, height=0.17\textwidth]{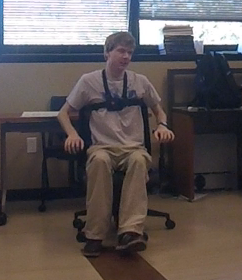}}\hspace*{0.1in}
  \subfigure[]{\includegraphics[width=0.36\textwidth, height=0.17\textwidth]{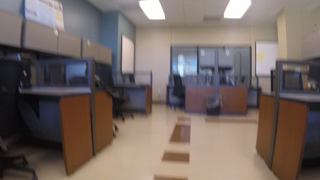}}\hspace*{0.1in}
  \subfigure[]{\includegraphics[width=0.144\textwidth, height=0.17\textwidth]{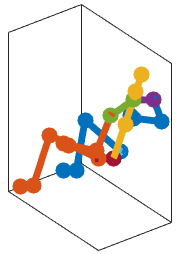}}
  \caption{Our goal is to infer the full 3D body pose of a person using the video captured from a single chest-mounted camera.
  (a): One person with a chest-mounted camera. (b): The egocentric view. (c): The predicted pose of the person using only video from view (b).
}
  \label{fig:concept}
\end{figure}



%

We ask the question: \emph{Is it possible to estimate the ``invisible" human body pose behind a single egocentric view?}  See Fig.~\ref{fig:concept}.
Despite the fact that we cannot see the person behind the body-mounted camera, the video seen from his point of view provides clues that may well be learnable.  In particular, we expect clues from two sources: \emph{dynamic motion signatures} and \emph{static scene structure}.  First, there exist scene-independent motion signatures for pose changes. For example, the act of standing up has a certain motion pattern as seen by the ego-camera, no matter if he stands up from a chair in a restaurant or a bench at the park.  In fact, first-person games use these effects to guide the virtual camera, giving gamers the impression they are moving the same way as the virtual character.  Second, static scene structure sets the context and offers a prior on likely poses.  For example, the pose of typing on a keyboard occurs in similar views showing a monitor or laptop, even though the hands need not be visible.  Or, if we see a table in front of us with a specific distance and angle, we can predict whether we are standing or sitting in front of the table.  

Of course, not all poses are distinguishable from egocentric video; some will be aliased, meaning different poses can produce the same visual signal.  Our intent is to leverage the typical structure linking how the scene changes to how the body is posed.

We introduce a novel approach to predict first-person body pose, given an egocentric video sequence.  As training data, our approach takes videos from a wearable camera, where each frame is labeled with ground truth pose parameters.  The pose is parameterized by 25 3D joint positions, i.e., a ``stick figure" representation, and is obtained with Kinect during training.  At test time, we are given a novel egocentric video from a new user, and must infer the sequence of 3D body poses based on the single wearable camera video alone.  

Our learning approach capitalizes on the clues described above, while also incorporating longer term pose dynamics.  First, classifiers based on dynamic and static cues estimate the probability of each of a (large) set of quantized poses per frame.  Then, we jointly infer poses for a longer sequence (1 to 3 minutes) based on those initial predictions together with a non-parametric model of pose dynamics.  The latter is used to identify a least-cost ``pose path" through exemplar training video.  This step regularizes the initial estimates with priors about how people can move, and is efficiently optimized with dynamic programming. The whole approach is fast---about 0.5 seconds per frame.

We validate our method quantitatively on videos from ten camera wearers performing daily activity poses, as well as qualitatively on  challenging videos in unconstrained environments.  The experiments show the proposed method gives robust results.  It greatly outperforms several alternative methods, including a CNN regression method modeled after the third-person DeepPose~\cite{deeppose} approach retrained for our setting.


In summary, our contributions are: (1) We tackle a new
problem that estimates the wearer's ``invisible" pose from a single ego-centric video;
(2) We propose a novel global optimization
method that leverages both learned dynamic and scene
classifiers and the pose coupling over a long time span;
and (3) We benchmark several methods, including hand crafted
features and CNN learned features, for our task.

\section{Related work}

We deal with a new problem of predicting invisible human poses from a single ego-centric video stream.  To put our idea in context, we review related work on third-person pose, egocentric hand/arm estimation, and egocentric activity analysis.

\paragraph{Third-person pose} Pose estimation from images and video has been studied for decades \cite{pose-survey-2015}.  Existing work tackles pose estimation from a third-person viewpoint, where the person is entirely visible.  In contrast, we consider estimating the body pose of the person \emph{behind} the camera; his/her body parts are rarely visible, if at all.  As such, existing pose estimation methods are not applicable to our scenario.

Some third-person pose methods use regression to map from images to pose parameters (e.g.,~\cite{deeppose,triggs,psh}), including the recent DeepPose work using convolutional neural networks~\cite{deeppose}.  At a glance, a direct regression approach seems like a possible solution for our problem.  Even though the body is not visible, we want to learn the connection between what the person sees and how his body is posed.  However, a naive application of that idea is inadequate, since 1) even large training sets cannot fully capture the possible variation in environments, poses, and movements, and 2) the relevant egocentric visual signals are inherently temporal.  The proposed method learns the connection between pose dynamic and static cues from snippets of video, and enforces long term constraints between estimated poses.  Our experiments show this yields superior results to a DeepPose-like scheme applied to our task.


\paragraph{First-person pose}
Limited research explores ways to infer the body pose of an egocentric camera wearer~\cite{deva,insideout,ren-gu,kitani-iccv2013,kitani-cvpr2013}.  Given interest in understanding handled objects, some methods are dedicated to estimating pixel-wise 2D maps of the camera wearer's hands~\cite{ren-gu,kitani-iccv2013,kitani-cvpr2013}. Recent work also investigates how depth data from an egocentric RGBD camera can help estimate shoulder, arm, and hand poses in 3D.  Both lines of work assume the body parts are visible in the egocentric view.  In contrast, we aim to estimate the full body pose of the person (e.g., 25 joint positions), and we do so even when the body is entirely out of view of the egocentric camera.

In this sense, our goal is more related to the ``inside-out" mocap approach of~\cite{insideout}.  In that work, 16 or more body-mounted cameras are placed on a person's joints, and then each camera's 3D location is recovered via structure from motion (SfM).  There are important differences with our technical approach and motivation.  First, rather than 16+ cameras attached at joints worn expressly for the purpose of a mocap session~\cite{insideout}, we employ a single chest-mounted camera---the sort typical wearable-computer-users may wear anyway while going about daily activities.  Thus, the SfM approach cannot be directly applied to our setting, and our system requirements are more lightweight and flexible.  Secondly, our approach is novel.  Whereas the mocap method employs a geometric solution to localize the joints, we devise a \emph{learning} solution that discovers the connection between how the ego-centered scene changes as a function of body pose.  We also note that the mocap method requires substantial computational resources---about 1.5 days for a minute of capture~\cite{insideout} due to SIFT matching expenses---whereas our method requires only 15 minutes.   The possible disadvantage of our method relative to~\cite{insideout} is our need for representative training data, though the data is relatively easy to collect, given that it requires no manual annotations (see Sec.~\ref{sec:dataset}).



\paragraph{Egocentric activity analysis}
Most recent egocentric vision work studies activity recognition~\cite{fathi,peleg,pirsiavash,farhadi,kitani-activity,spriggs,yinli} or object recognition~\cite{damen,ren-gu}.  Once again, the focus is largely on visible activity happening in front of the camera---particularly hand-object manipulation activities.  However, some work shows that ego-actions (like riding a bus, snowboarding, etc.) are detectable from the scene video~\cite{kitani-activity,peleg}, and the walking style of the camera wearer can even aid person identification~\cite{style}. We consider whether ego-video can go further to reveal full 3D body pose.  While we also use movement information, our method does not infer action classes.   For instance, rather than recognize the current action as ``walking", our approach will produce the detailed pose across the walking cycle.  Thus, our method provides a mid-level representation---explicit pose---which could be further used in high-level activity recognition or other applications.

%


%

\section{Method} \label{sec:method}

We estimate 3D human poses
from the video of a chest-mounted camera.
Predicting human poses from egocentric video is essentially a regression problem: from the input video,
we estimate the 3D position of each body joint in the wear's local frame.
In the following sections, we give details about instantaneous pose estimation using local features
and full sequence estimation using the pose path method.

\subsection{Pose parameterization and data collection} \label{sec:dataset}

We use a Kinect V2 sensor
to capture the ground truth human poses.
Pose is represented as the 3D positions of 25 body joints defined in the MS Kinect SDK.
The predicted 3D pose is positioned in a local coordinate system.
The first axis is parallel to the ground and points to the wearer's
facing direction. The second one is parallel to the ground and in the same plane as the shoulder line.
The third axis is perpendicular
to the ground. The joint coordinates are normalized by five times
the shoulder length of the subject. 

In data capture, the subject wears a chest-mounted camera to record egocentric video.  
We choose chest-mounted (vs.~head-mounted) because they provide a stable view unaffected by constant head bobbles.  
The frame rate of both the Kinect sensor and the ego-camera is 30Hz.
The two are synchronized
using time stamps.
We capture a total of 18 ground truth videos, in which 3 videos are for training and the rest
for testing.
Ten subjects with different height, body shape, and gender are involved
in data collection.
They are instructed to perform normal daily activities in public places
such as offices, labs, and libraries.
Our ground truth dataset is collected indoors due to the limitation of
Kinect V2 sensor.
However, our approach is general, and we demonstrate outdoor tests as well. With more advanced motion
capture setups, our method can be trained
in even broader action domains.  

\subsection{Instantaneous pose estimation}

We construct a function $f(v,p)$
that gives the probability of video segment $v$
corresponding to pose $p \in P$, where $P$ is the set
of all possible poses. In this paper, $P$ includes all the poses in 
the train sequence.
Here $v$ is a mini-sequence of egocentric video frames, e.g. a one-second clip. In the following,
we also use $v$ to represent the feature vector extracted from a video
segment.
Due to the large number
of possible poses, directly constructing  $f$ is difficult.
We therefore introduce pose clusters as an intermediate pose representation.
We cluster the normalized poses using $k$-means with $L_2$ norm to obtain
$K$ pose clusters.
For everyday movement, $K=300$ is sufficient. Then we train a classifier to obtain the function $g(v,c)$ to extract the probability of
video segment $v$ matching the pose cluster $c$.
The mapping $f$ is approximated as $f(v,p)=g(v,c(p))$,
where $c(p)$ is the pose cluster identity of a pose $p$.

\subsubsection{Dynamic clues}
Egocentric video has specific motion patterns for different human
movements. Human poses thus have strong correlation
with the scene dynamics in the egocentric view. This is more so for
the transient poses.
A human
observer can often infer the wearer's pose from the global
scene motion.
\begin{figure}
\centering
\includegraphics[width=\linewidth]{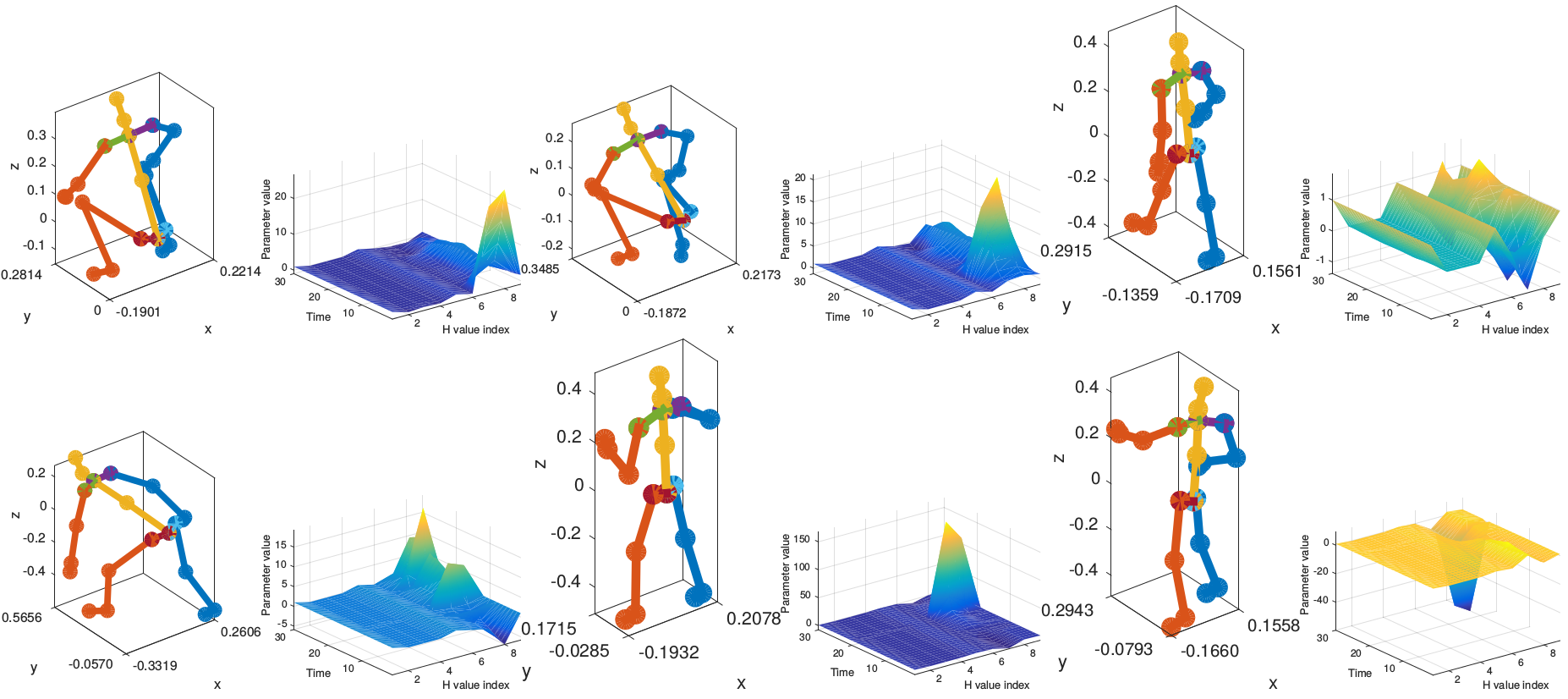}%
\caption{
Example poses and the corresponding dynamic features for the surrounding 1-second video segment.  Similar poses often have similar dynamic features (see first two examples), and distinct poses have different features.
}
\label{fig:dyna-feature}
\end{figure}

To construct a feature that is scene invariant,
we extract the sequence of homographies between successive
video frames. Strictly speaking, the homography is scene invariant only when the camera is purely
rotating.  However, the egocentric camera translates very little
between successive frames if the frame rate is high, making the
camera rotation dominant and the representation close to scene invariant.
This useful property allows us to use very few
training data to obtain good classifiers 
(as opposed to attempting to learn appearance-specific cues, which would be overly restrictive to a given training environment.)
  The feature is still related
to the camera's intrinsic matrices. If the camera matrix $K$ is known, we
can normalize the result by computing the approximate rotation matrix $R=K^{-1}HK$, where $H$ is
the homography.

To compute a homography between frames, we use optical flow to find the point correspondence. 
A least squares method is used to estimate the homographies,
which can be implemented using SVD. The elements in each homography
are then normalized by the top-left corner element. The stack of normalized
homographies
over a fixed time interval (one second), is used to represent
the camera movement. Fig.~\ref{fig:dyna-feature} illustrates how the proposed feature helps 
differentiate poses of the wearer.  In Fig.~\ref{fig:dyna-feature},
the homographies are vectorized and combined into a matrix in each one-second time interval.

Using the above feature,
we train a random forest to predict the probability of the pose at each instant of the input video
belonging to each of the $K$ pose clusters.
We build 100 random trees with arbitrary depth.
The dynamic feature classifier gives reasonable results. However, the result is
ambiguous when there is little motion in the egocentric
video. 
To resolve this issue, we also use static scene
structure, as defined next.


\subsubsection{Static scene structure clues}
\begin{figure}[t]
\centering
\includegraphics[width=0.8\linewidth]{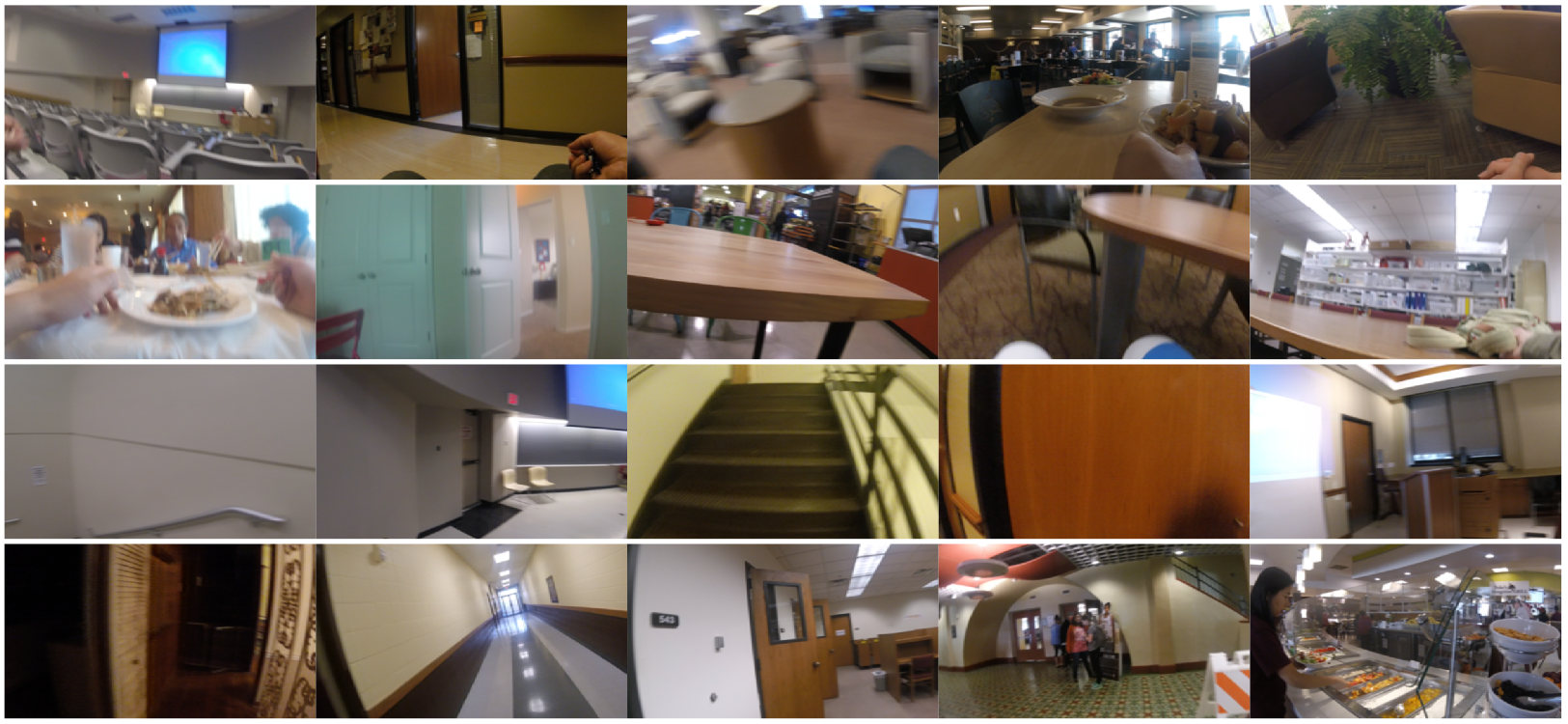}
\caption{Samples from the training dataset of sitting (Rows 1-2) and standing (Rows 3-4).}
\label{fig:sitstand}
\end{figure}

In everyday life, two static poses are most common: standing and sitting.
Many dynamic poses are often similar to these broad categories, e.g., walking is standing-like 
and kneeling is sitting-like.   Indeed, in the dataset in \cite{peleg}, roughly $95\%$ of frames 
can be classified as standing-like or sitting-like.
\footnote{We do not include lying down because it does not happen often in the day.}

Given image $n$ from the egocentric view, we compute $h_n$, the probability  the corresponding
pose
is sitting-like; its probability to be standing-like is $1-h_n$.
We collect a training dataset containing 5,530 standing images
and 2,946 sitting images in different indoor environments.
Fig.~\ref{fig:sitstand} shows sample images from the dataset.
We train a CNN classifier
by fine tuning the last three layers of the fully connected network in
the AlexNet \cite{alexnet}; the learning rates of other layers are set to be zero.
The two-class classifier generalizes well.
On our ground truth dataset with 71,623 egocentric video frames and
poses from Kinect V2,
the sitting-like and standing-like image classification accuracy is $65.09\%$ and $77.97\%$, respectively.
The dataset is composed of $79.71\%$ images with standing-like poses.

%

\subsubsection{Local cost of pose estimation}

Thus far we have provided two ways to estimate pose for each frame, using dynamic and static cues.  
These instantaneous estimates are not the final output of our system, however.  As we will explain 
in Sec.~\ref{sec:nonparam}, errors can be corrected in a global optimization stage where we infer 
the entire \emph{pose path} over the entire sequence.

In particular, the two classification outputs above serve as unary terms of an energy function for 
the longer sequence of surrounding frames (1-3 minutes per clip in our dataset).  Let $x_{i,n}$ be an indicator
variable, which is 1 if at time $n$ the pose $i$ is predicted. Here $i$
is the id of a pose
in $P$.
Let $e_{i,n}$ be the cost of predicting pose $i$ at time $n$. The
overall unary cost term is $U = \sum_{n=1..N, i \in P} e_{i,n} x_{i,n}$,
where $N$ is the number of frames.
Here $P$
is also used to represent the ids of all the possible poses.
We define the cost $e_{i,n}=1-g(v_n,c(p_i)) + d_{i,n}$, where $g$ is the probability
of dynamic feature $v_n$ being classified to pose cluster $c(p_i)$, and
$d_{i,n}$ is determined by both the static scene classifier and the dynamic scene classifier.
We use $d_{i,n}$ to penalize the selection of pose $i$ at time $n$
if there is large chance that the estimations from
the dynamic features and the static features mismatch.
Specifically, we define $d$ as:  
$d_{i,n}=\delta$ if
$h_n > \tau$ and $\hat g(v_n)$ is standing, or $h_n < 1-\tau$ and $\hat g(v_n)$ is sitting,
and otherwise 0. Recall that $h_n$ is the probability of sitting from the
static scene feature at time $n$.
The $\hat g$ is derived from $g$ to classify the current video frame as sitting or standing using the dynamic features and random forest, based on the known listing of which pose clusters are sitting/standing-like.

Simply optimizing the local cost is not sufficient. Without considering
the inter-frame pose constraints the pose predictions can be noisy. Another issue is the resolution.  Since the local pose cost is estimated from
the probability of quantized poses, it tends to
be a staircase function over time.
In the following, we show how to solve both of these problems by optimizing
poses simultaneously over a long time span.

\subsection{Non-parametric prior on pose dynamics}\label{sec:nonparam}

Next we show how we optimize the final sequence of pose estimates based on the local costs and a 
non-parametric prior on pose dynamics.  First we define the prior, then we introduce an efficient optimization approach.

\subsubsection{Pose paths in an implicit motion graph}
To infer a likely sequence of poses over time, our method constructs an implicit motion graph
that controls the possible transitions between  poses in the exemplar training videos.
The graph nodes correspond to poses in exemplar videos. The edges indicate possible transitions
from one pose to another.


The optimal pose sequence corresponds to the optimal \emph{pose path} on an exemplar pose sequence.
The pose path is composed of a sequence of ``steps'', each of which represents a transition
from one pose to the next.
We enforce that each step
  can only move from a pose cluster to the same pose cluster or a direct neighbor pose cluster.
Each pose in the exemplar pose sequence belongs to a pose cluster.
We define pose clusters as direct neighbors if we can find two poses that are
drawn from each of the two pose clusters and are  
adjacent in time in the exemplar pose sequence. 
Since the same
pose cluster may appear at different times in the exemplar pose sequence, the above rule
allows large jumps. To further regularize the pose path, we constrain the step sizes, uniformity
of the step sizes, and control the stationary steps on the pose path (see below).
Therefore when determining where a step should lead to, we also have to consider
previous decisions on the pose path. Thus, the transition costs
dynamically change with the traversal history.

This graph is reminiscent of motion-graphs used for motion synthesis in
computer graphics~\cite{gleicher,motiongraph}.  However, whereas motion synthesis aims
to generate convincing movements within an annotated mocap database based on a few user-specified
anchor poses, our task is to jointly
infer the sequence of poses in a novel egocentric video. Furthermore, unlike traditional motion graphs, 
edge weights in our graph dynamically change to allow the regularizers mentioned above.


We have used $P$ to denote the set of all the poses in the training dataset.
Here we overload the notion; we also use $P$ to represent the concatenation of all the
training pose sequences from the training dataset. The poses in $P$ thus 
preserve the original temporal order.
Selecting a sequence of poses from $P$ is equivalent to
find a path on $P$ so that the following energy function
is minimized:
\begin{align*}
& \min_X \{U(X)+T(X)+V(X)+S(X)\} \\
& \mbox{s.t. } \mbox{The assignment of } X \mbox{ represents a sequence of poses drawn from } P.
\end{align*}

Here $X$ is the matrix $[x_{i,n}]$, where $n$ is the time index and $i$ is the index of poses in $P$ and
recall that $x_{i,n}$ is a binary variable to indicate whether pose $i$ is selected
at time $n$. To represent a path, at each time instant $n$, we have $\sum_i x_{i,n} = 1$.  
Here $U(.)$ is the unary term defined in the previous section.
$T(.), V(.), S(.)$ are terms that control coupling between poses in the whole sequence.
$T(.)$ constrains the step size between successive footprints
on the path, $V(.)$ controls the speed of the pose transition, and
$S(.)$ restricts stationary steps.

\subsubsection{The step size term $T$}

We order the pose ids in $P$ according to their temporal
sequence in the training video.
If we choose pose $l \in P$ at instant $n-1$, we say we step on point $l$ at time $n-1$.
At time $n$, we may step to $l+k$, where $k$ is the step size from time $n-1$
to $n$. Since the original exemplar video is continuous, the smaller the $k$
the smoother the pose transition is likely to be. If the step size is $0$,
we keep the same pose in the time interval. The stationary step can be used to infer a
slower movement in the testing video. If the step size is 1, the movement
has the same speed in the training and testing video. For $k > 1$, the movement
in the testing video is faster than the exemplar sequence.
In the energy function, we prefer the step size to be small and at the same
time we allow occasional large jumps from one point to the other.

In particular, $T= \sum_{i,j,n} w_{j,i} x_{j,n-1}x_{i,n}$, where $w_{j,i}=0$
if $i-j\le 2, i \ge j$ and otherwise $w_{j,i}=\delta$, where $\delta$ is a positive constant penalizing backward steps
and steps that are great than two. 
Apart from the step size constraint,
we also constrain that if $c(p_i) \ne c(p_j)$ and $c(p_i)$ and $c(p_j)$ are
not consecutive in the training video $w_{j,i}=+\infty$. Here $c(p_i)$ is the pose cluster of pose $i$.
This prohibits the
path from going from one pose to another with too much difference or using a transition
of pose clusters not seen in the exemplars.  However, it does
allow long jumps from one pose cluster to the same pose cluster or one that is
a direct neighbor to the cluster. However, such long jumps do have a penalty. So, we prefer that steps on the path
move to a directly adjacent frame if possible.  We allow the
path to go forward or backward. 

\subsubsection{The speed smoothness of the path $V$}

The above step size term roughly enforces a first order constraint on the
path: small steps are taken when possible. However, the path
may still have a non-uniform speed of steps in a short time span, which is undesirable
because within a time of 1 or 2 seconds human body motion is usually uniform. We thus introduce a
second order term to penalize the speed changes:
\[
    V = \sum_{i,j,n} q(|s_{j,n-1}-(i-j)|) x_{j,n-1}x_{i,n} \;,
\]
where $s_{j,n-1}$ is the speed at time $n-1$, for step $j$.
Here $q$ is a truncated linear function: $q(x)=\mu x$ if $x < \gamma $ and
otherwise $q(x)=\mu \gamma$, where $\gamma$ and $\mu$ are constant parameters.
This term encourages the path to maintain a constant speed.

\subsubsection{The stationary step penalty $S$ in the path }
Simply minimizing the first order and second order smoothness of the path
is not enough. Recall that the local cost in short time intervals tends to be constant.
The steps in the pose path thus tend to be stationary because the first
and second order smoothness terms will be zero. The step size penalty helps but
is not sufficient. We thus penalize stationary steps:
\[
    S = \sum_{i,j,n} r(u(j,n-1),i) x_{j,n-1}x_{i,n} \; ,
 \]
where $r(u(j,n-1),i)=0$ if $i \ne j$, otherwise $r(u(j,n-1),i)=t(u(j,n-1)+1)$.
We therefore count the number of stationary steps and penalize
the pose stop changing for a long time.
Here, $u(i,n)$ is the number of stationary steps accumulated at
time $n$ if the current pose is $i$; $u(j,n-1)$ is similarly defined.
Similar to $q$, $t(.)$ is a truncated linear function.
The stationary step penalty term
thus makes the path less likely to stay at one point and helps
resolve the temporal resolution loss problem.

\subsubsection{Optimizing the pose path using DP}
We can rewrite the problem into a recursion:
\small
\begin{align*}
& H(i,n) = e_{i,n}+\min_{j \in S_i}\{H(j,n-1) + w_{j,i} + q(|s(j,n-1)-(i-j)|) + r(u(j,n-1),i)\} \\
& u(i,n) = u(j^*,n-1) + 1, \mbox{if }j^*=i\mbox{ and otherwise } u(i,n) = 0 \\
& s(i,n) = i - j^*, \; \; p(i,n) = j^*,
\end{align*}
\normalsize
where \small $j^* = arg\min_{j \in S_i} \{H(j,n-1) + w_{j,i} + q(|s(j,n-1)-(i-j)|) + r(u(j,n-1),i)\}$.\normalsize

\noindent $S_i$ is the set of poses that can transform to $i$. Here, $H(i,n)$ is the optimal energy of pose path
if the path ends at a specific pose $i$ at time $n$.
$u(i,n),s(i,n),p(i,n)$ are the stationary step number, speed of steps and
previous optimal pose selection of the optimal pose path ending at pose $i$ at time $n$.
We initialize $H(i,1)= e_{i,1}$, $u(i,1)=0,s(i,1)=0, \forall i \in P$. All the other $H$ are initialized to be $+\infty$,
and $p$ to be $-1$.
We can verify that solving the recursion is
equivalent to optimizing $\min_X \{U(X)+T(X)+V(X)+S(X)\}$, where the solution of $X$
represents the pose path.
The recursion can be efficiently solved using dynamic programming (DP).

\begin{figure}
\framebox{
\subfigure[]{
\includegraphics[width=0.7\textwidth]{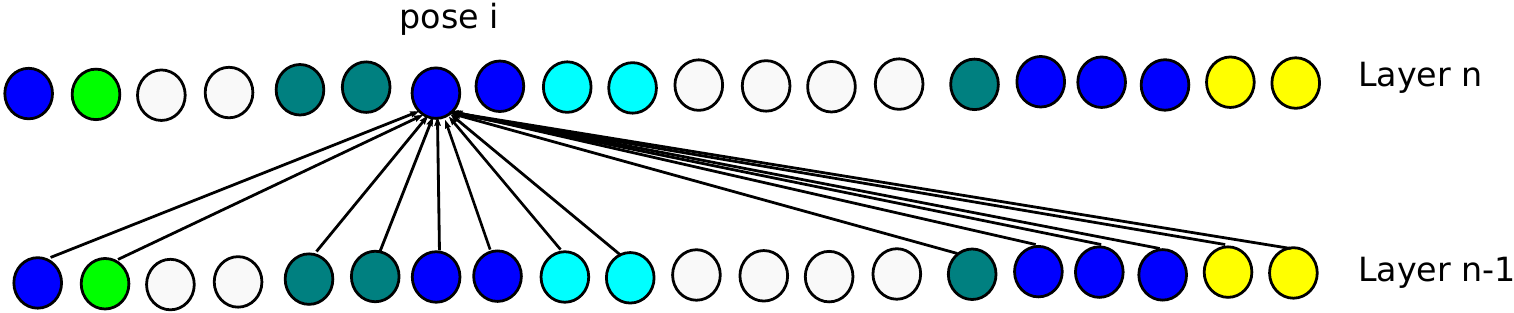}}}%
\framebox{
\subfigure[]{
\includegraphics[width=0.195\textwidth]{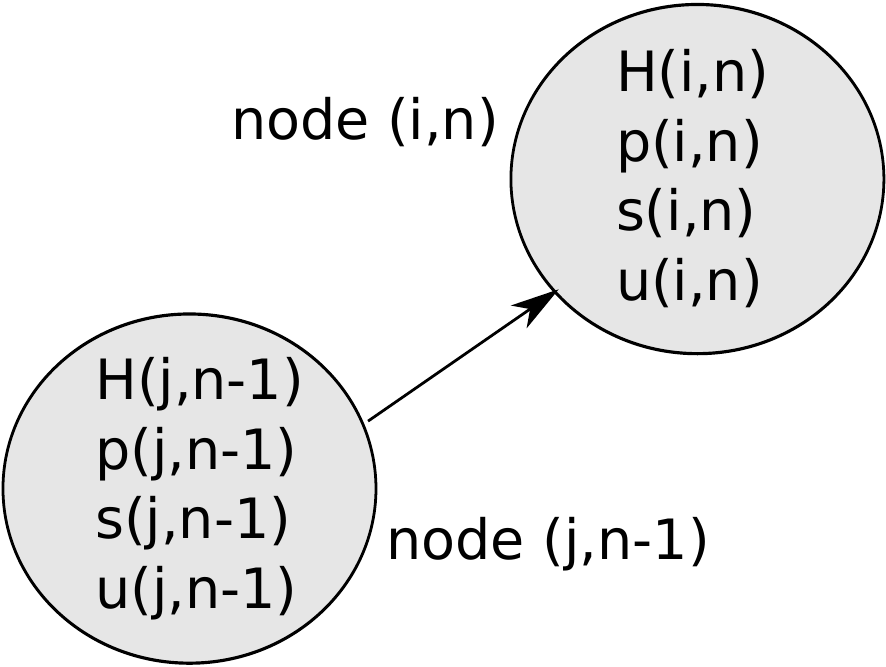}}}
\caption{(a): Connection between two trellis layers.  Colors indicate
pose clusters. We only allow pose transitions to the same or neighboring
 cluster. In this example, the blue cluster's neighbors have colors: light green, dark green, light blue and yellow. 
(b): State variables in each node.}
\label{fig:trellis}
\end{figure}

It helps to visualize the optimization in a trellis.
The trellis contains $M$ columns and $N$ rows, where $M$ is the number of possible poses in $P$
and $N$ is the number of input video frames. Fig.~\ref{fig:trellis} illustrates the edge connection
from layer $(n-1)$ to node $i$ in layer $n$.
Each edge corresponds to one possible step in the path.
Each node has a cost $e_{i,n}$, where $i$ is the column and $n$ is the row of the trellis.
Each edge has a weight $w_{j,i} + q(|s_{j,n-1}-(i-j)|) + r(u(j,n-1),i)$. The DP finds a
minimum cost path in the trellis.

Solving the DP involves updating the state variables $H,s,u,p$ in each node. Since only the nodes inside
the same or neighboring cluster are connected by each stage of the trellis, the
complexity is much lower than $O(M^2N)$. Moreover, we can use the local pose probability to prune impossible nodes
from the trellis. In fact, most of the poses have near zero probability from the random forest classifier. If we
only keep nodes that correspond to poses that have probability greater than 0.01,
the trellis becomes very sparse and the
corresponding DP can be quickly completed (typically contributing 0.01 seconds per frame for our whole system).
\section{Experimentation}

We evaluate the performance of the proposed method on both a ground truth dataset
and challenging videos in unconstrained environments.
(See videos at \url{www.cs.bc.edu/~hjiang/egopose/index.html}).

In our ground truth data, the 3D human poses are captured from the Kinect V2 for ten human subjects. 
The synchronized egocentric video
is from a chest-mounted GoPro camera.
Below we consider two settings.  In the first setting, 
training and testing videos are from the same human subject, but taken in disjoint indoor environments such as lab, office, hallway and living room.
In the second setting,
the training and testing videos are from different human subjects
\emph{and} recorded at different locations.  
There are in total 71,623 test video frames (about 40 minutes) in the ground truth experiments, consisting of clips ranging from 1-3 minutes each. 
We also test about 15 minutes of video from unconstrained video, 
which lacks ground truth for evaluation.



\paragraph{Implementation details}

For the unary term $U$, we set $\delta=0.1, \tau=0.99$.  We thus include a penalty $\delta$ only when the confidence of the sitting-standing classifier
is above $99\%$.
For the truncated linear functions $q$ and $t$, we fix $\gamma=10, \mu=0.01$ and $\gamma=5, \mu=0.02$, respectively.
All parameters were set based on manual inspection of a few examples during method development, then fixed for all experiments.  
With sufficient labeled data, their values could be set with DP to minimize pose errors.

\renewcommand{\tabcolsep}{1pt}

\begin{figure}[tb]
 \centering
 \includegraphics[width=\linewidth]{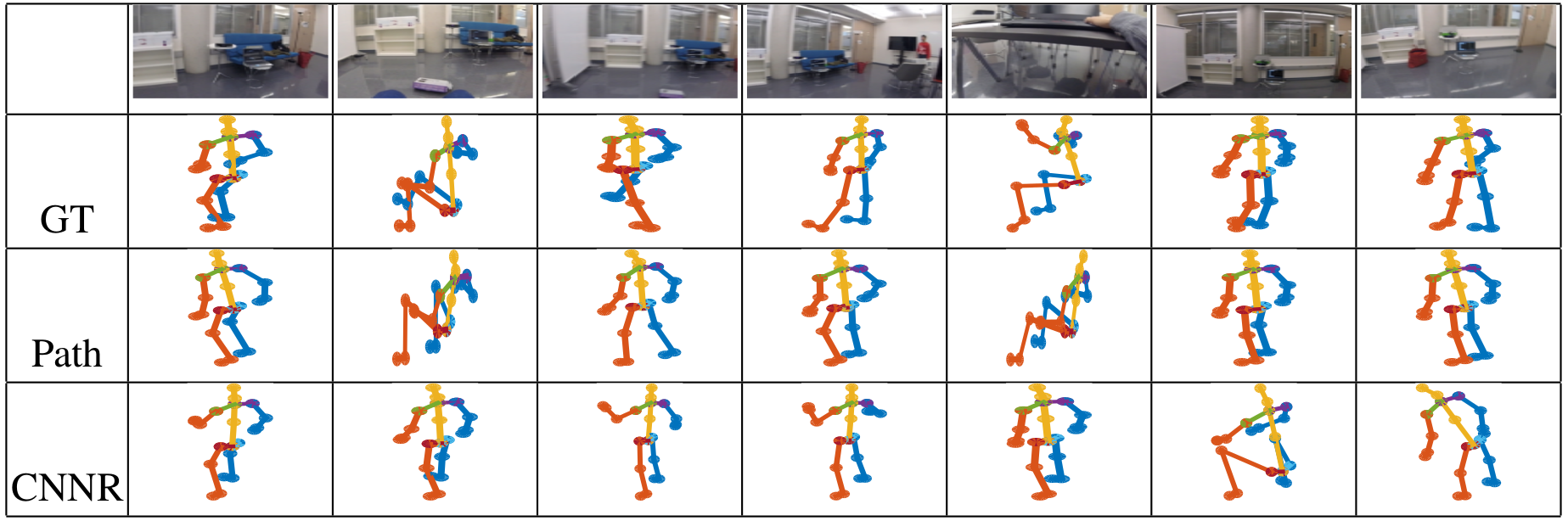}
  \caption{Comparison with the DeepPose \cite{deeppose} method retrained for our task. GT: ground truth. Path: proposed method.
   CNNR: \texttt{CNN-Regression} baseline.}
  \label{fig:dif}
\end{figure}

\paragraph{Baselines}

No prior work predicts body pose from egocentric video. We therefore devise a series of informative baselines to gauge the impact of our method, including methods inspired by today's best image-based third-person pose estimators:

\begin{itemize}
\item \textbf{\texttt{CNN-Regression}}: an adaption of the DeepPose \cite{deeppose} method to our task. Our problem is still a regression problem,
even though the camera wearer is not visible from the egocentric view. We use the same network structure as the
DeepPose except that our input is a stack of grayscale images in every one-second video clip and output is the 25 body joints defined by the Kinect SDK.  We scale each image to 100 $\times$ 100.  

\item \textbf{\texttt{KdTree}}: simple nearest neighbor approach using Kd-trees.  It finds the ``closest'' video segments in the training data and then
takes the corresponding 3D poses as the prediction result.
The stacked homography in every 30 frames is used
as the feature, and the $L_2$ norm is the distance metric. Other norms give similar results.  

\begin{figure}[t]
\centering
\includegraphics[width=0.8\linewidth]{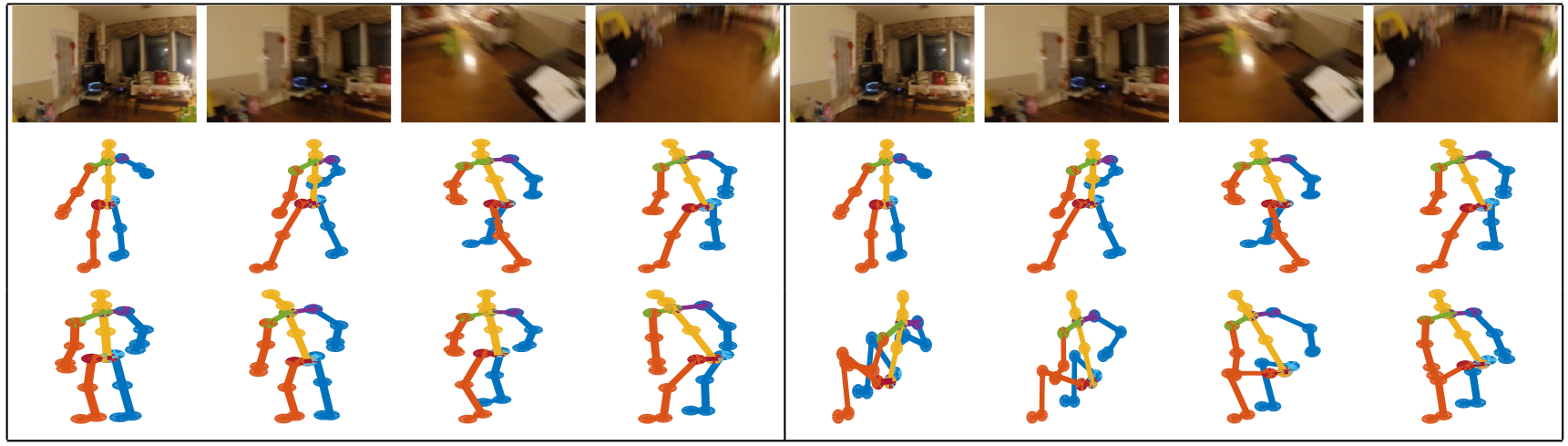}
\caption{Comparing with the \texttt{Kd-tree} baseline. Row 1: Sample frames. Row 2: Ground truth poses.
Row 3: Our result in left box and Kd-tree result in right box.}
\label{fig:kdtree}
\end{figure}

\item \textbf{\texttt{Path-Cluster}}: a variant of the proposed method.
Instead of directly optimizing the poses, this method first finds the pose clusters and then refines the pose estimates using dynamic programming.
The refinement is similar to the proposed pose path optimization, except that the pose candidates at each instant can only come from the pose clusters estimated in the first stage.

\begin{figure}[t]
\centering
\framebox{
\includegraphics[width=0.1\textwidth]{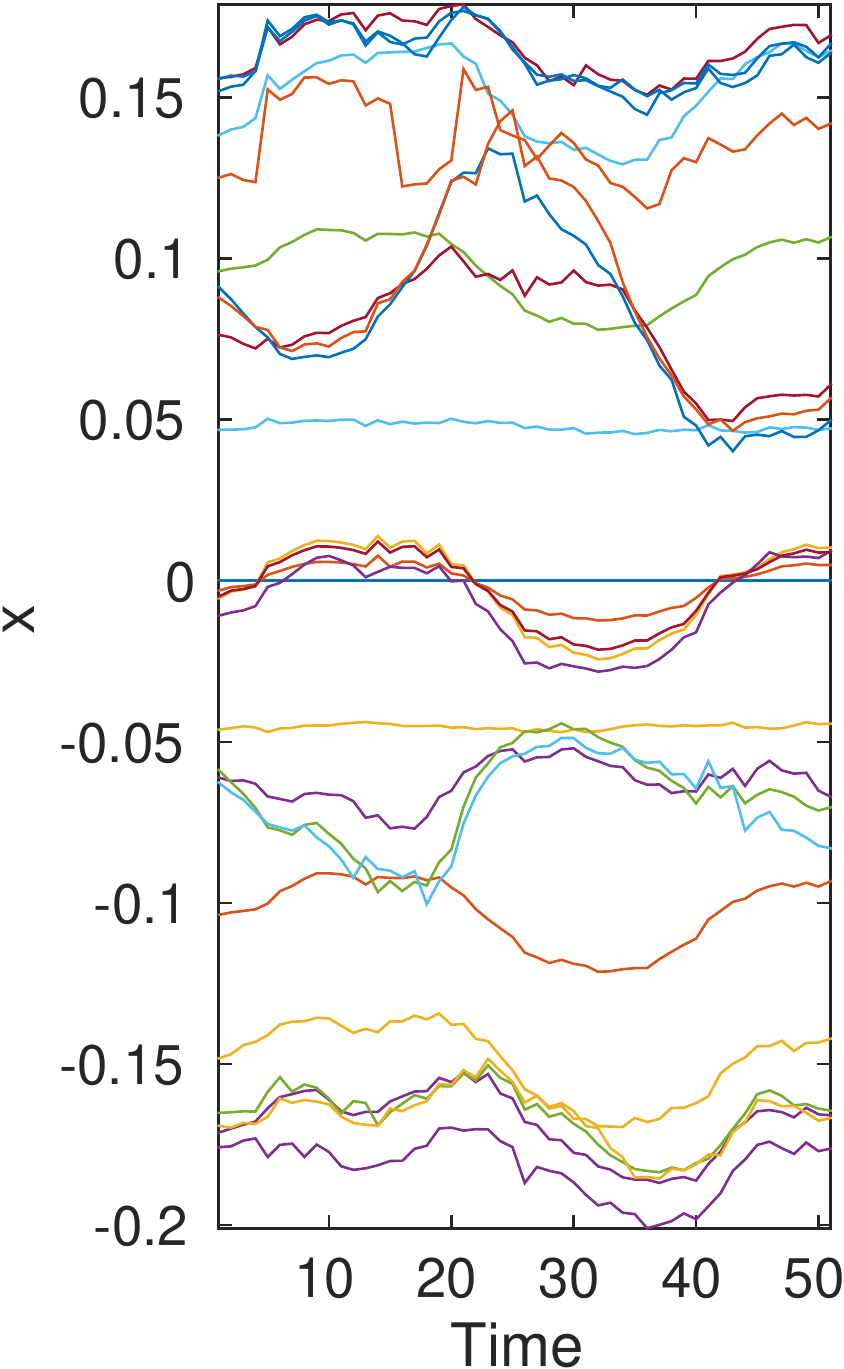}%
\includegraphics[width=0.1\textwidth]{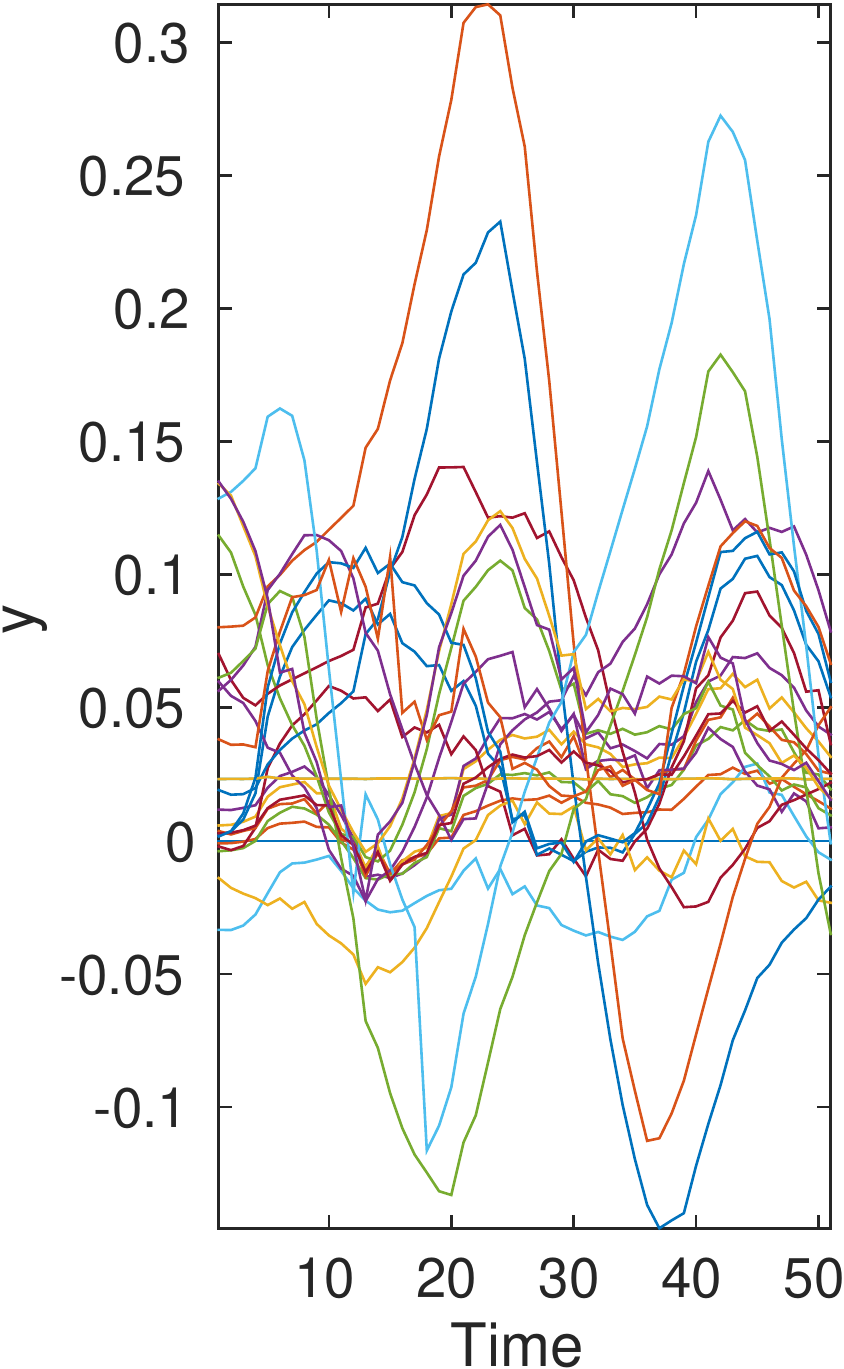}%
\includegraphics[width=0.1\textwidth]{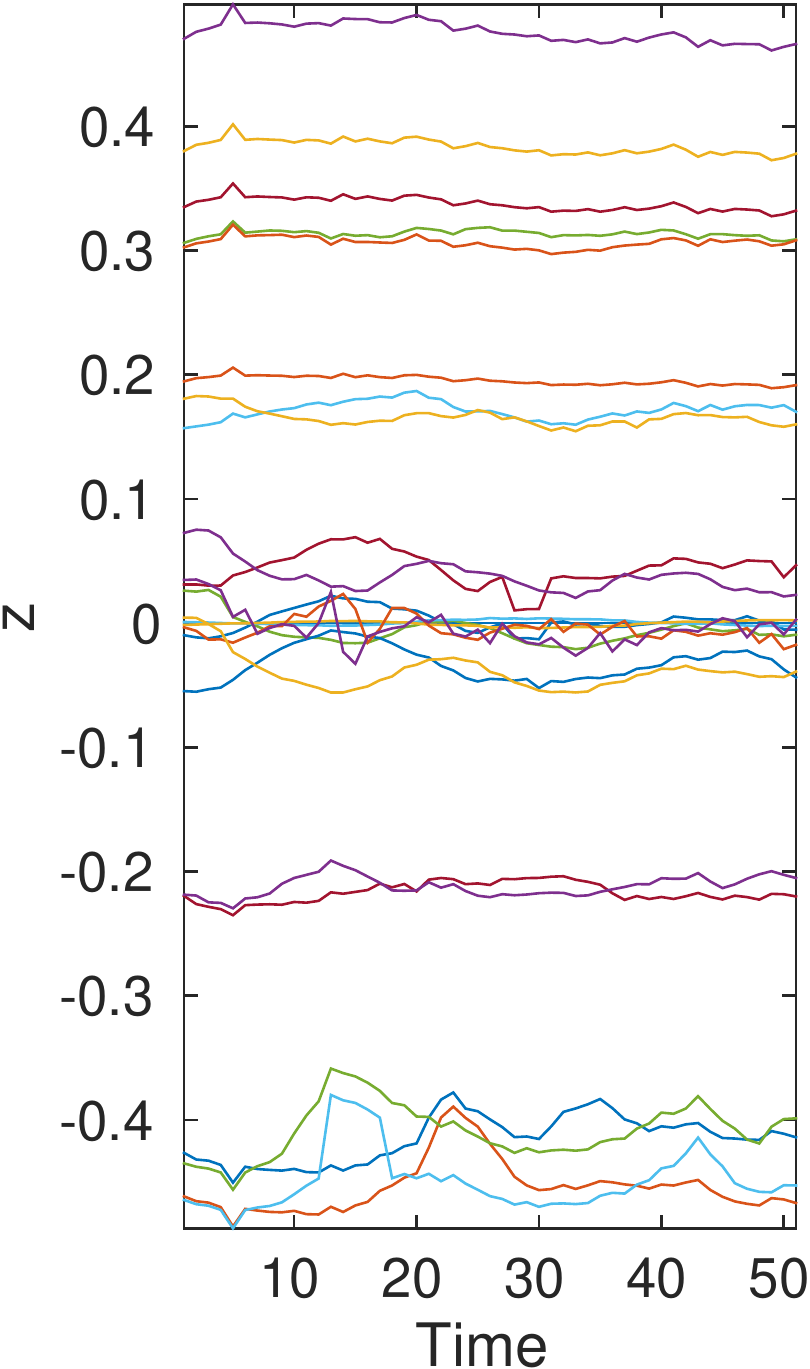}%
}%
\framebox{
\includegraphics[width=0.1\textwidth]{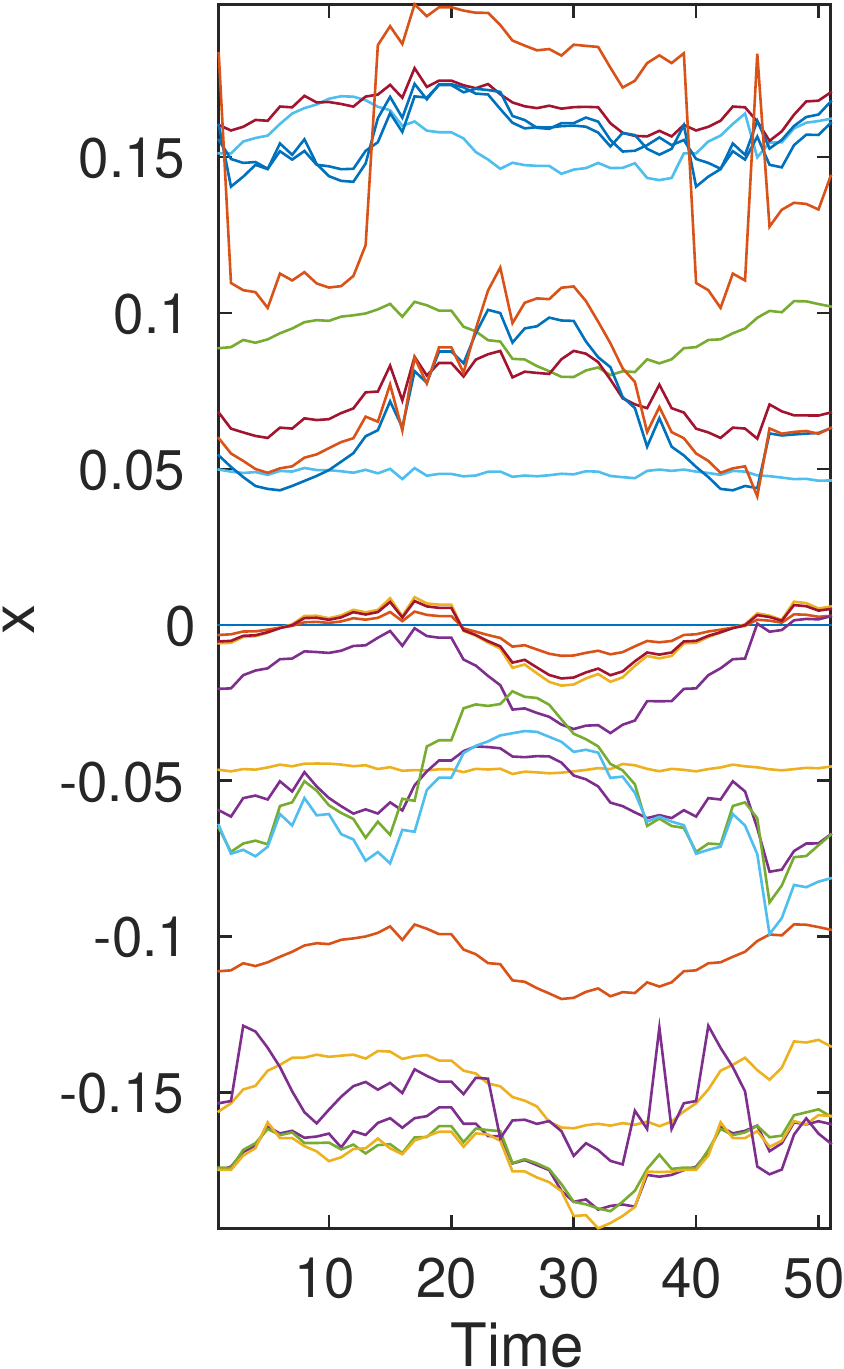}%
\includegraphics[width=0.1\textwidth]{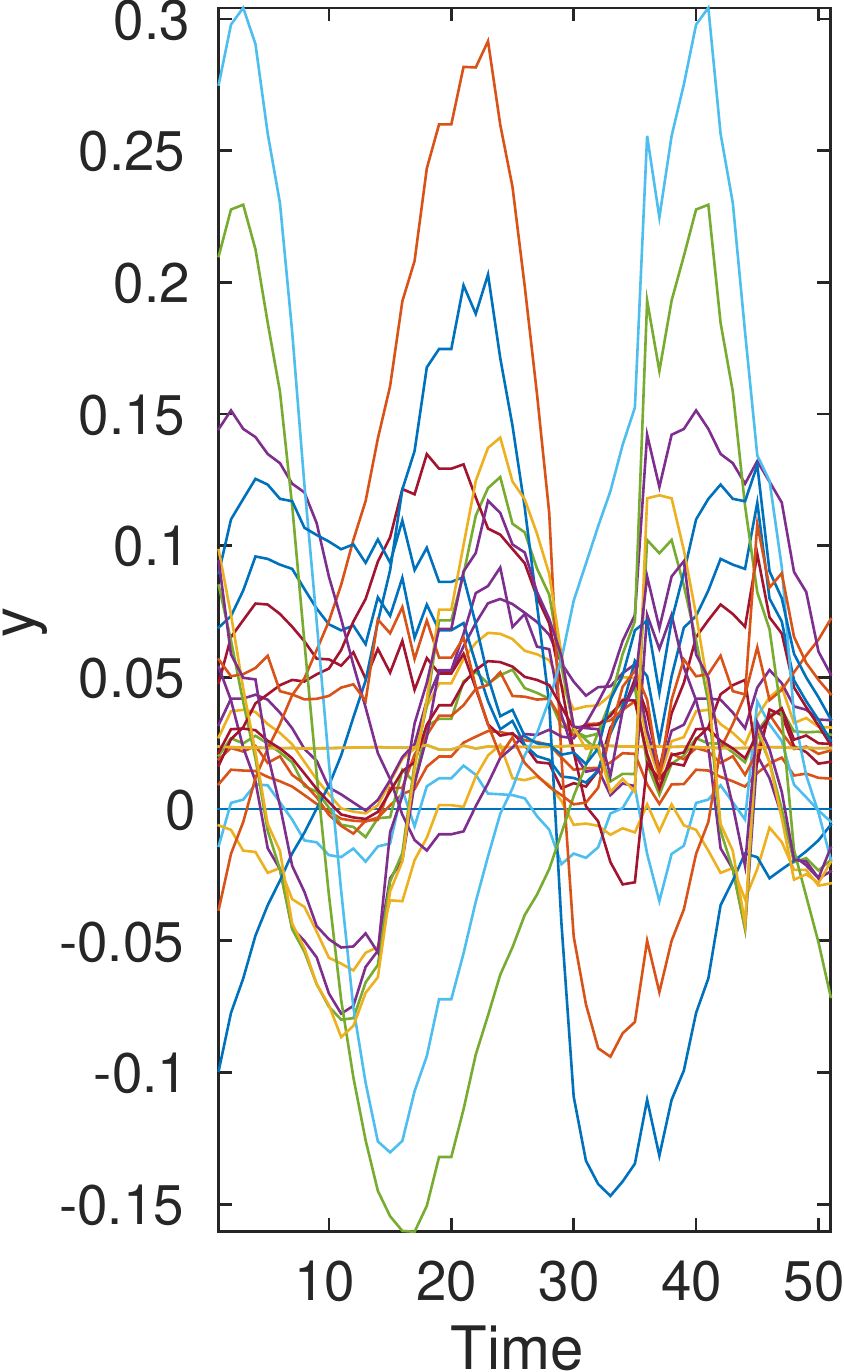}%
\includegraphics[width=0.1\textwidth]{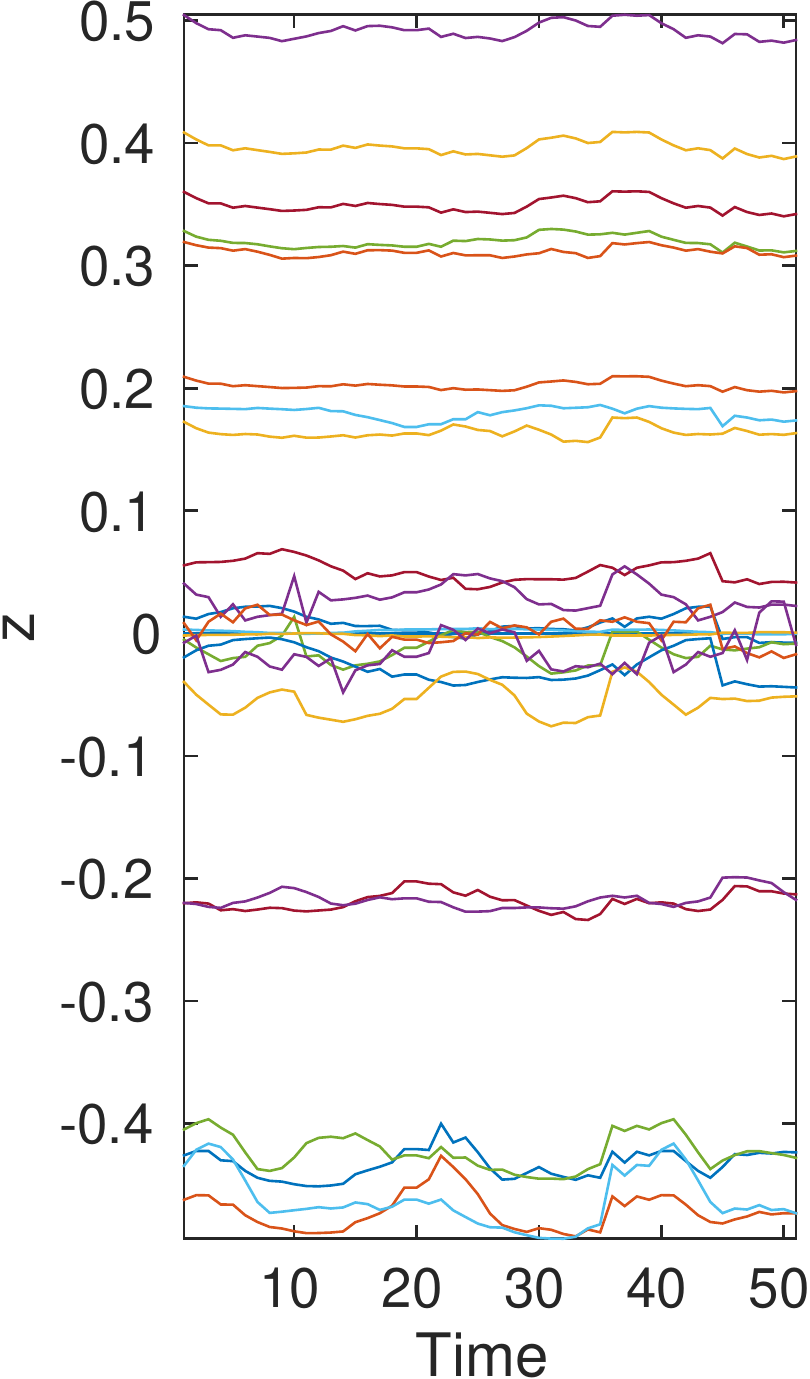}%
}%
\framebox{
\includegraphics[width=0.1\textwidth]{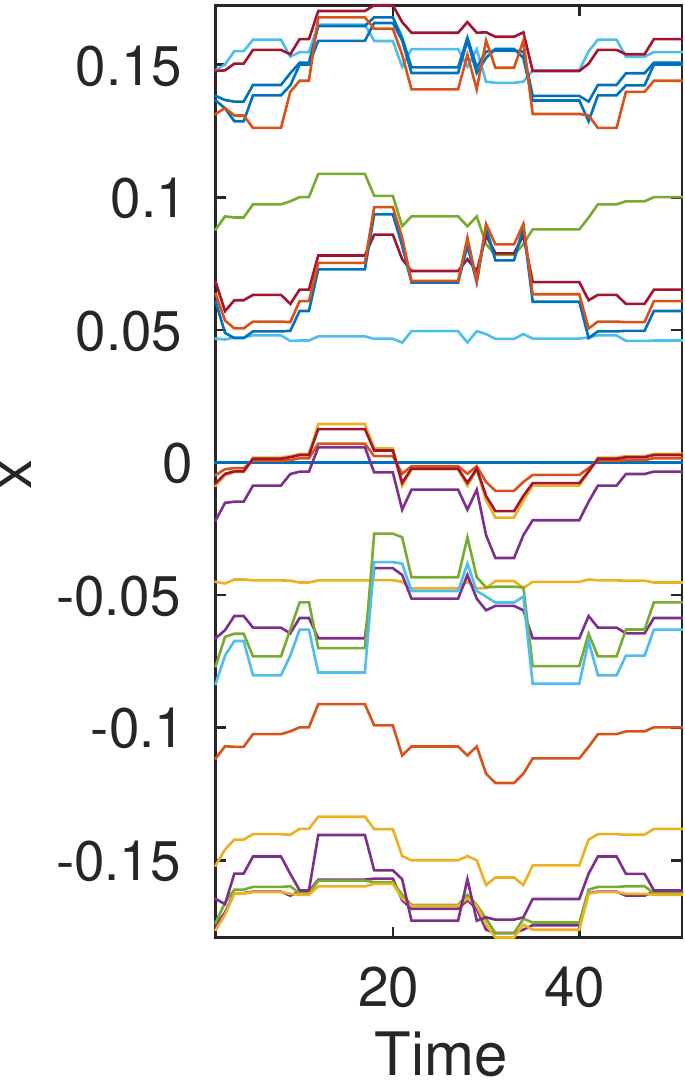}%
\includegraphics[width=0.1\textwidth]{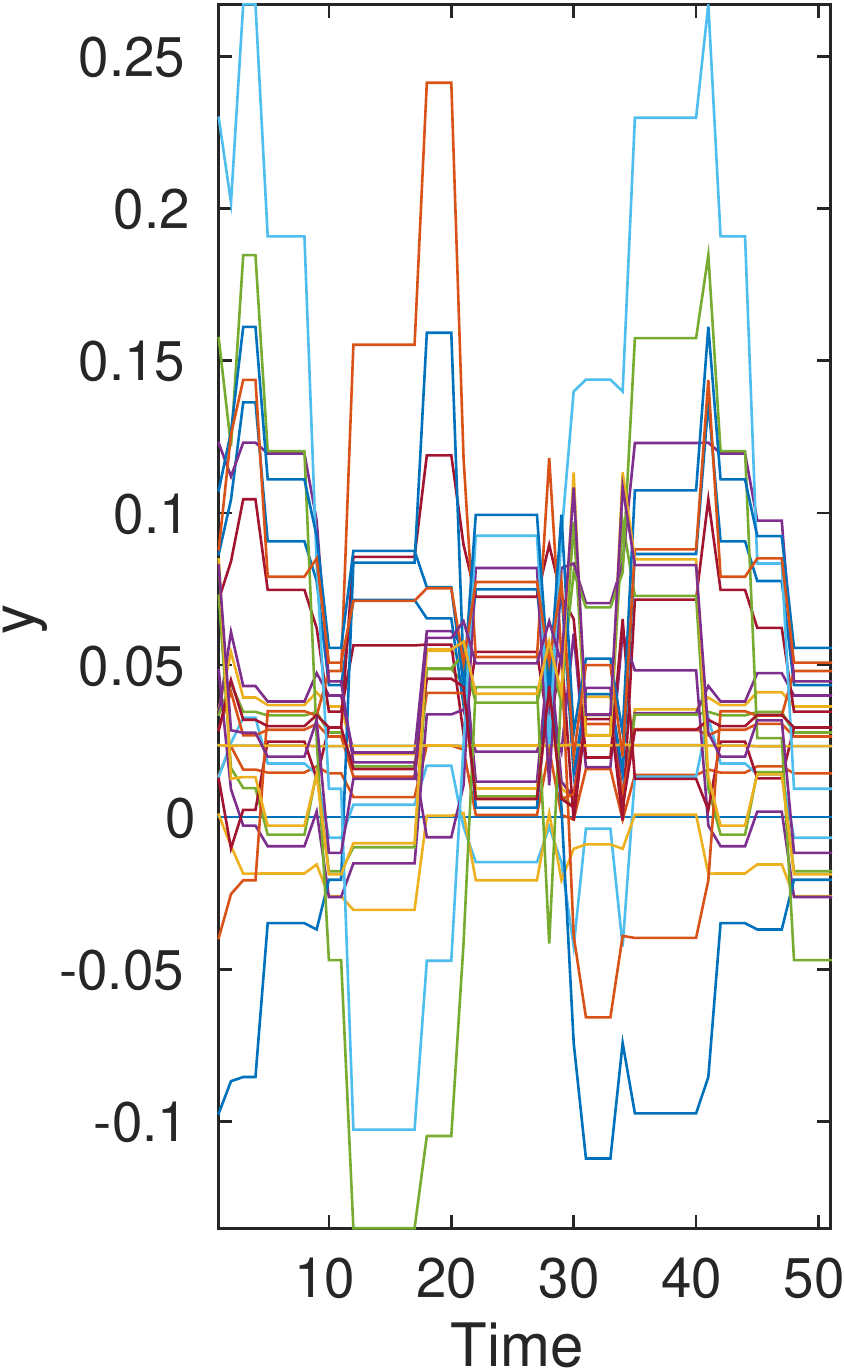}%
\includegraphics[width=0.1\textwidth]{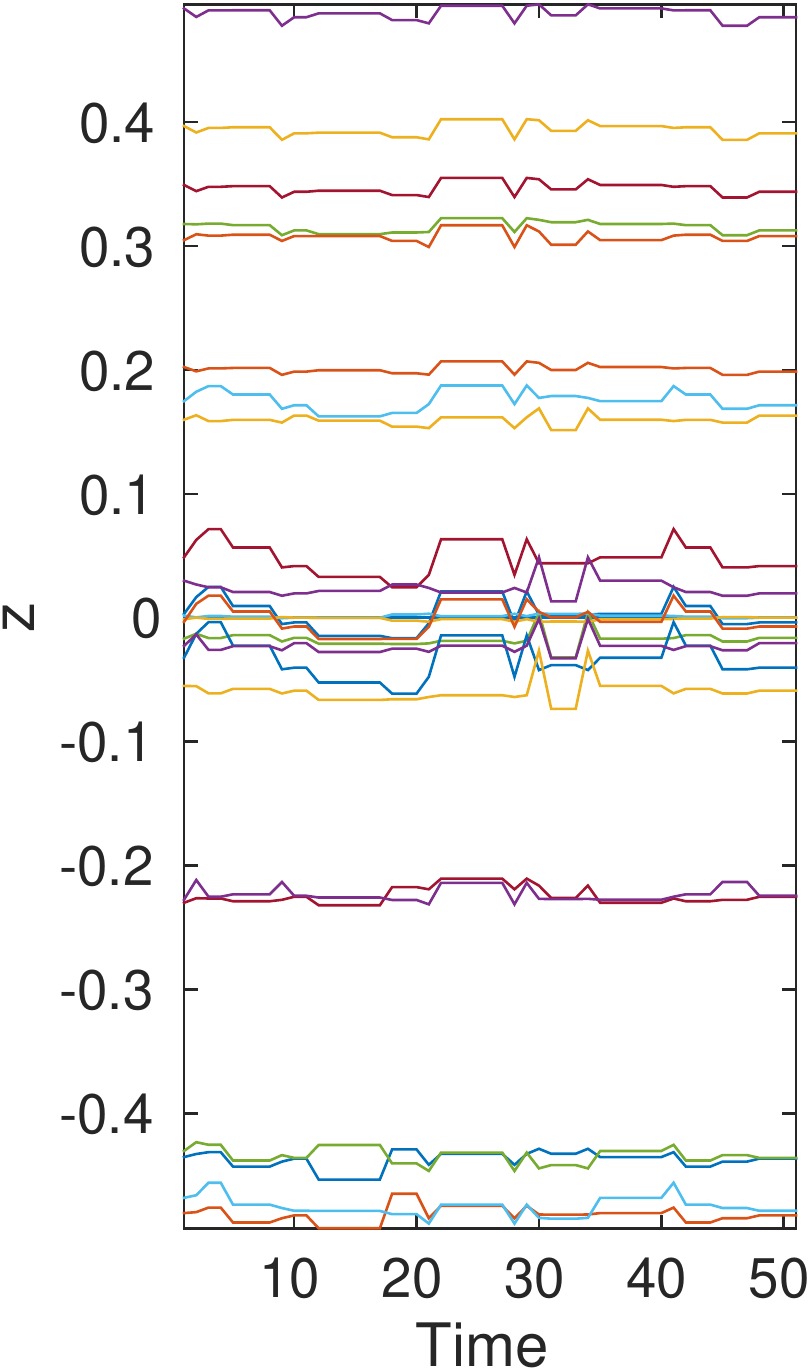}%
}%
\caption{
The $x,y,z$ coordinates of ground truth body joints (left), versus the results from proposed method (middle) and pose cluster centers of \texttt{Path-Cluster} (right) on a 50-frame video.
}
\label{fig:posecluster}
\end{figure}

\item \textbf{\texttt{Path-CNN}} and \textbf{\texttt{Path-CNN-Refine}}:
a variant of the proposed method that uses deep-trained features in place of our hand crafted homography features.  We train a deep neural network to classify each sequence of 30 frames to one of the 300 pose clusters. We use AlexNet \cite{alexnet} due to its good results in many applications.
In the first setting (\texttt{Path-CNN}), we rescale each input video frame to 100 $\times$ 100 and retrain the network from scratch.
In the second setting (\texttt{Path-CNN-Refine}), we fine-tune on the modified AlexNet with depth 30. The fine-tuning is only on the
first convolution layer and the last three fully connected layers. We compute the local pose cost as one minus
the class probability from the CNN output. The proposed global optimization is then applied to obtain the final result.


\item \textbf{\texttt{AlwaysStanding}} and \textbf{\texttt{AlwaysSitting}}: simple guessing methods that exploit the prior that
 poses are typically somewhere near a standing or sitting pose (hence much stronger than a truly random guess).  
We compute the standing and sitting poses by the average over training subjects.
    
\end{itemize}

\begin{figure}[tb]
\includegraphics[width=\linewidth]{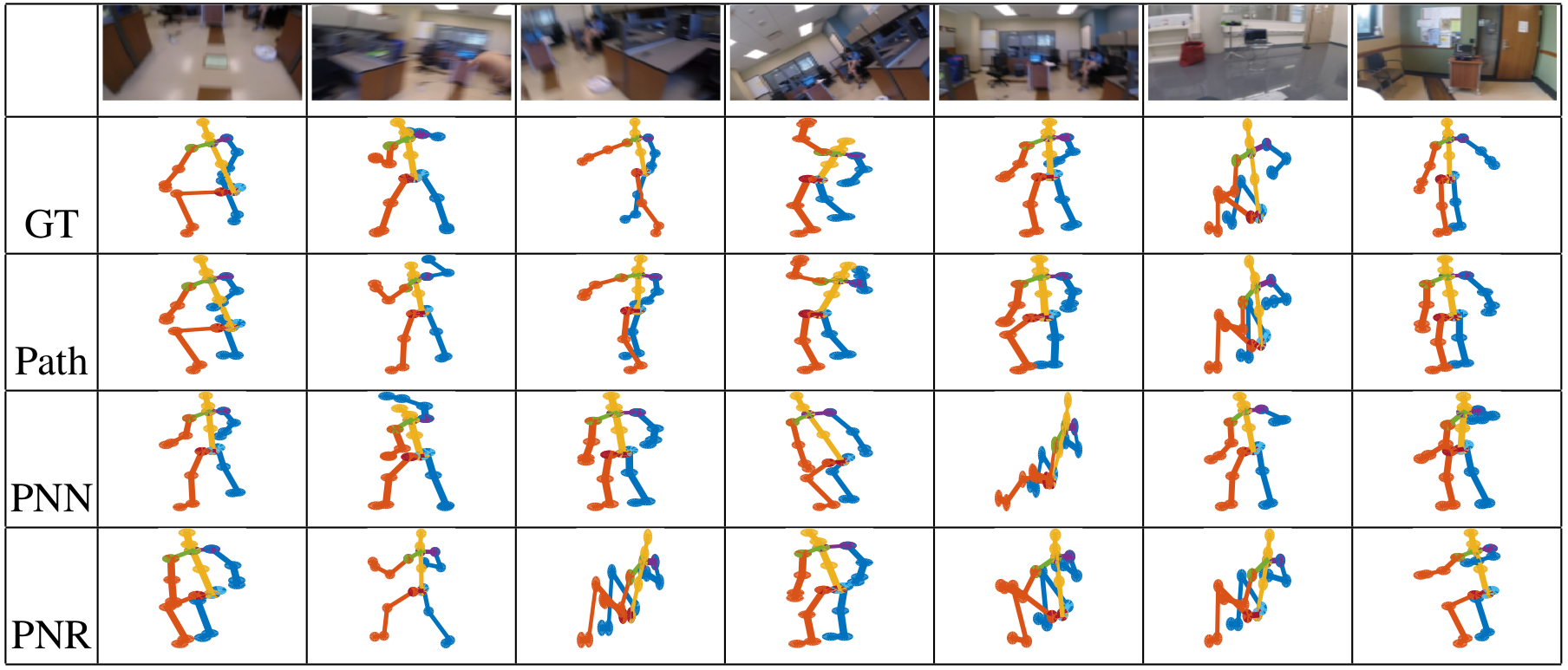}
  \caption{Comparison with methods using deeply learned features. GT: ground truth. Path: proposed method. PNN: \texttt{Path-CNN}. PNR: \texttt{Path-CNN-Refine}.}
  \label{fig:same}
\end{figure}

Figs.~\ref{fig:dif} and \ref{fig:kdtree} show qualitatively that the proposed method indeed gives better results than the DeepPose adaptation (\texttt{CNN-Regression})
and nearest neighbors (\texttt{KdTree}), neither of which considers long term pose coupling.
Our method also gives better results than the three variations
of the proposed method as shown in Figs.~\ref{fig:posecluster} and \ref{fig:same}.

Fig.~\ref{fig:posecluster} shows that if we directly use the the estimated pose cluster centers
as the predicted poses, the results have lower temporal
resolution than the proposed method. 
Refining the pose selection in each estimated pose cluster
is inferior to the proposed approach because the errors in the first stage cannot be undone.
The predicted pose sequence is also not
as smooth as the proposed method.
\texttt{Path-Cluster} is essentially
an interpolation method that smooths the cluster centers estimated in the first step.
Note that a simpler linear interpolation method is not directly usable because it
does not always give valid poses.

Fig.~\ref{fig:same} shows qualitatively that using deep neural networks to train the dynamic and scene structure features does not give better results. Neither training from scratch nor fine-tuning improves the result.
Neural network approaches need a large dataset to capture different variations of the scene and human poses.
Our method is able to train on a small dataset
and achieve good performance.

Now we present the quantitative comparisons with all baselines.  
We analyze the errors of the joints with highest variance in everyday activity: head, elbows, wrists, knees, and ankles.
In the wearer's local frame joints such as shoulders and hips do not vary much in normal daily activities.  
Fingers and toes are also not included because they are not accurately estimated by Kinect;
they mostly follow the wrists and ankles.  
We quantify error by the distance between the predicted 3D joints
and the ground truth, after rotating each joint point cloud so that the shoulder is parallel to the $yz$ plane and
the body center is at the origin.  Recall that the predicted coordinates are already in normalized coordinates according to the shoulder length of the subject.  We convert raw errors to centimeters based on a reference shoulder joint distance of 30 cm.

Tables~\ref{tab1} and \ref{tab2} show the results, for the two settings defined above.  Overall the proposed method gives
smaller errors than all the competing methods.

\begin{table}[t]
\centering
\scriptsize
\begin{tabular}{c|c|c|c|c|c|c|c|c}
\hline
  & Path (Ours) & Path-Cluster & Path-CNN & Path-CNN-R & KdTree & CNN-Regr & AwaysStanding & AwaysSitting\\
\hline
Head &15.8(0.08) &16.5(0.08) &21.6(0.14) &22.9(0.14) &18.1(0.11) &16.2(0.10) &\textbf{15.1}(0.08) &32.5(0.09)\\
Elbow &\textbf{14.4}(0.07) &15.4(0.07) &18.6(0.12) &19.4(0.12) &15.8(0.10) &14.4(0.09) &14.5(0.08) &20.7(0.08)\\
Wrist &\textbf{19.1}(0.09) &20.6(0.10) &26.5(0.17) &27.1(0.17) &21.3(0.13) &22.0(0.14) &22.9(0.12) &21.3(0.08)\\
Knee &\textbf{15.4}(0.09) &17.2(0.09) &27.3(0.17) &26.2(0.17) &22.0(0.14) &21.3(0.13) &21.2(0.11) &40.0(0.11)\\
Ankle &\textbf{20.7}(0.10) &22.9(0.10) &33.8(0.21) &33.3(0.21) &28.4(0.18) &26.4(0.17) &26.7(0.13) &37.9(0.09)\\
\hline
NAvgAll  & \textbf{17.2} &  19.1 &   48.1 &   48.7 &   32.8 &   29.7 &   24.6 &   31.9 \\
NAvg(W+A) & \textbf{19.9} & 22.6 &   60.0 &   60.2 &   40.8 &   38.7 &   32.4 &   27.1 \\
\hline
\end{tabular}
\caption{Average joint error (cm) and standard errors, when training and testing on same subject but in different environments.  The training sequence has 6,950 frames. There are 7 test videos with a total of 25,195 frames.
We compute the mean error normalized by the standard error for the nine joints denoted NAvgAll, and
for the wrists and ankles denoted NAvg(W+A).  
}
\label{tab1}
\end{table}

\begin{table}[b]
\centering
\scriptsize
\begin{tabular}{c|c|c|c|c|c|c|c|c}
\hline
  & Path (Ours) & Path-Cluster & Path-CNN & Path-CNN-R & KdTree & CNN-Regr & AwaysStanding & AwaysSitting \\
\hline
Head &16.6(0.07) &18.0(0.07) &19.4(0.09) &21.3(0.10) &20.1(0.09) &15.8(0.07) &\textbf{14.3}(0.07) &29.1(0.07)\\
Elbow &15.3(0.06) &16.9(0.06) &19.1(0.09) &19.5(0.09) &18.0(0.08) &15.8(0.07) &\textbf{14.9}(0.06) &20.9(0.06)\\
Wrist &\textbf{22.2}(0.08) &24.2(0.08) &29.7(0.14) &29.4(0.14) &24.9(0.12) &24.3(0.11) &23.8(0.09) &22.9(0.07)\\
Knee &\textbf{18.9}(0.07) &24.4(0.09) &21.6(0.10) &21.8(0.10) &31.9(0.15) &27.6(0.13) &21.7(0.08) &45.7(0.09)\\
Ankle &\textbf{24.9}(0.09) &29.9(0.10) &29.2(0.14) &29.2(0.14) &38.1(0.18) &33.3(0.15) &28.2(0.10) &43.0(0.09)\\
\hline
NAvgAll & \textbf{19.9} &   24.6 &   35.4 &   36.4 &   44.5 &   34.6 &   22.4 &   32.9 \\
NAvg(W+A) & \textbf{23.6} &  28.4 &  46.6 &  46.3 &  53.3 &  44.6 &  28.9 &  30.7 \\
\hline
\end{tabular}
\caption{Average joint errors (cm) and standard errors when training and testing on disjoint people and environments.  The training sequence has 10,000 frames from two subjects. There are 8 test videos with a total of 46,428 frames.
See Table~\ref{tab1} for the definition of NAvgAll and NAvg(W+A).
}
\label{tab2}
\end{table}

\begin{figure}[t]
  \includegraphics[width=\linewidth]{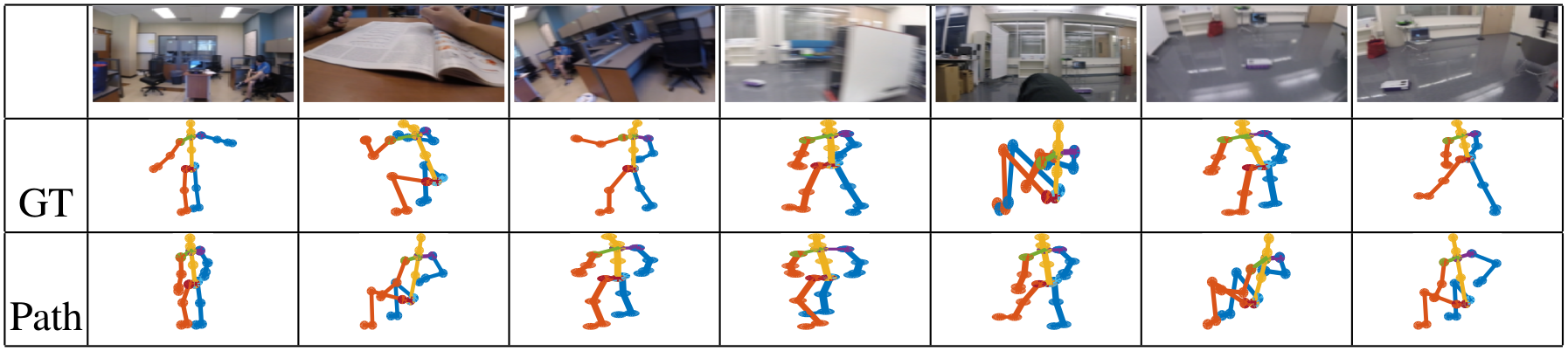}
  \caption{Failure cases.  See text for details.}
  \label{fig:fail}
\end{figure}

\begin{figure}[t]
  \includegraphics[width=\linewidth]{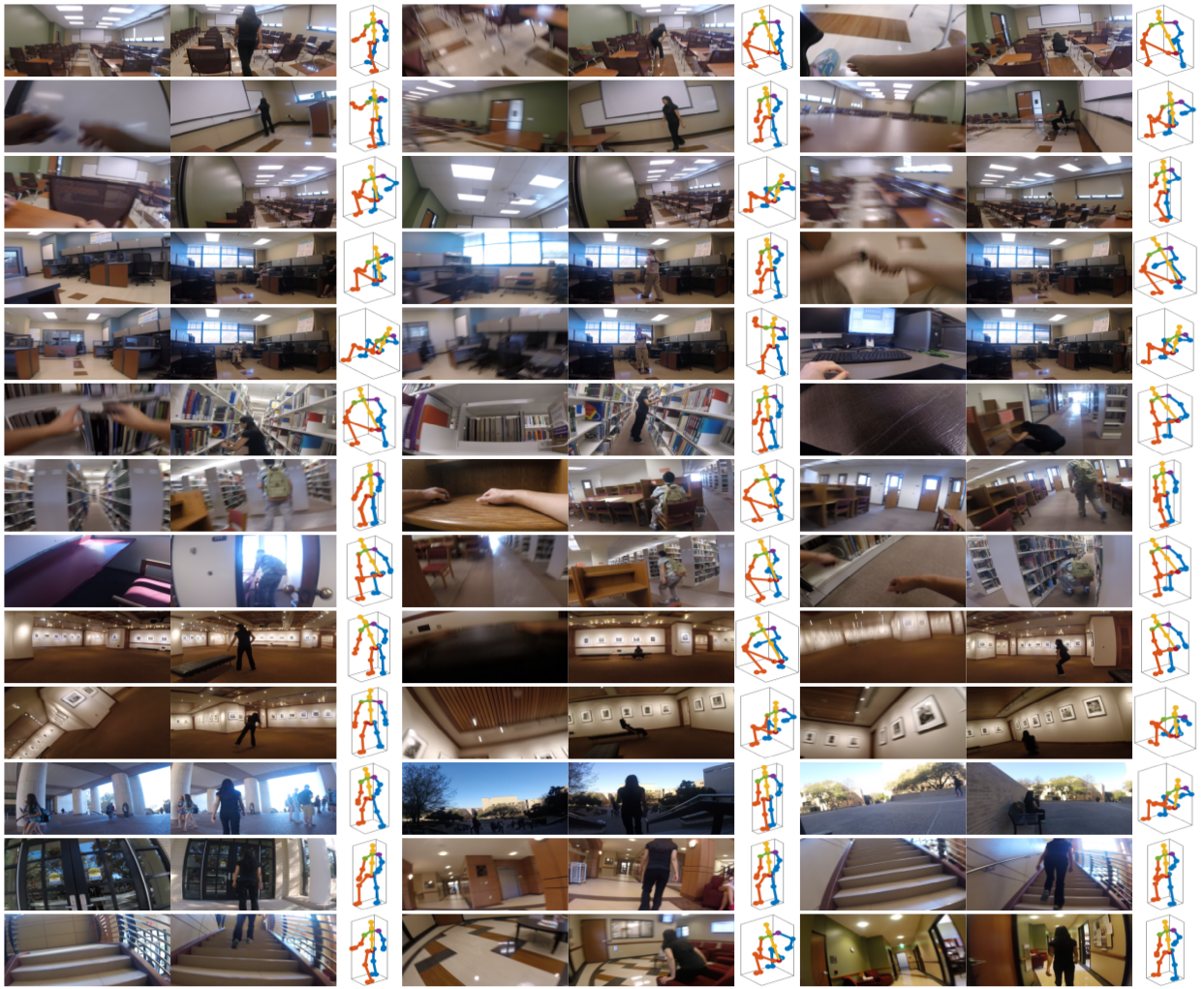}
  \caption{Experiments on data without ground truth. There are three subjects (S1, S2, and S3).
Row 1-2: Classroom (S1).
Row 3: Classroom (S2). Row 4-5: Lab (S3). Row 6: Library (S1).
Row 7-8: Library (S2). Row 9-10: Art gallery (S1). Row 11: Outdoor (S1).
Row 12-13: Hallway (S1). Each result contains three columns: egocentric view, side view (unseen by our method) and pose prediction.
}
  \label{fig:noground}
\end{figure}


The tables show our method outperforms the DeepPose-like \texttt{CNN-Regression}, presenting 
the value in our scene-invariant dynamic homography features.  It is also better than all the variants of our method we tested.  While \texttt{AlwaysStanding} is a reasonable prior for most test frames, our method still makes noticeable gains on it, showing our ability to make fine-grained estimates (e.g., 6 cm better on average for the ankles and knees). \texttt{AlwaysSitting} has much larger errors than any method, in line with the distribution of the test data.  Finally, between Table~\ref{tab1} and~\ref{tab2}, as expected we see that absolute error is lower for all methods with the benefit of observing the same subject during training.  



Fig.~\ref{fig:fail} shows some failure cases by our method.
Failures are mostly due to the ambiguity of the input.  Arm poses are not always predictable, if they do not affect the motion or the viewing angle of the egocentric camera.  
Other errors are due to the misclassification between the standing
and sitting poses.
Improving the features and instantaneous
pose estimation accuracy may further improve the results.

Finally, we test our method on 8 video sequences with no ground truth, captured in varying environments and with 3 subjects.\footnote{It is easy to capture egocentric video in arbitrary environments for test data; it is the Kinect ground truth capture for training that places restrictions.}
The training dataset is the same as above.  Fig.~\ref{fig:noground} shows sample results.
For each example, we display the frame from the egocentric camera as well as one from a side camera viewing the subject.  Note that the side view is for display only, and never used by our method. The 3D pose is estimated using only the egocentric video.  
Our method works well on this data, including for outdoor test sequences despite all training taking place indoors.  Please see the Supp video.

\section{Conclusion}
We tackle a new problem in computer vision: predicting human poses from egocentric video.
The proposed global optimization method is able to give accurate pose predictions in both same-person  and cross-person tests.  
Our experiments show our method gives superior results to a number of alternative approaches. We believe our method will be useful for many different applications including
egocentric video logging, summarization, and information retrieval, and it could facilitate action and movement understanding.

\section{Acknowledgments}
This research is supported in part by U.S. NSF 1018641 (HJ)
and ONR PECASE N00014-15-1-2291 and a gift from Intel (KG).


\end{document}